\documentclass[a4paper,fleqn]{cas-dc}

\usepackage[authoryear]{natbib}

\usepackage{algorithm}
\usepackage{algorithmic}
\usepackage{subcaption}
\usepackage{booktabs}
\usepackage{multirow}
\usepackage{amsmath} 
\usepackage{amsfonts}
\usepackage{amssymb}
\usepackage{float} 

\def\tsc#1{\csdef{#1}{\textsc{\lowercase{#1}}\xspace}}
\tsc{WGM}
\tsc{QE}
\tsc{EP}
\tsc{PMS}
\tsc{BEC}
\tsc{DE}

\begin{document}
\let\WriteBookmarks\relax
\def\floatpagepagefraction{1}
\def\textpagefraction{.001}
\shorttitle{Neural Networks}
\shortauthors{Kun Zou et~al.}

\title [mode = title]{GAFD-CC: Global-Aware Feature Decoupling with Confidence Calibration for OOD Detection}

\author[1]{Kun Zou}
\ead{zouk7@mail2.sysu.edu.cn}
\credit{Writing - original draft, Writing – review \& editing, Investigation, Validation, Methodology, Conceptualization}

\author[1]{Yongheng Xu}
\ead{xuyh59@mail2.sysu.edu.cn}
\credit{Writing – review \& editing}

\author[2]{Jianxing Yu}
\ead{yujx26@mail.sysu.edu.cn}
\credit{Supervision}

\author[1]{Yan Pan}
\ead{panyan5@mail.sysu.edu.cn}
\credit{Supervision}

\author[2]{Jian Yin}
\ead{issjyin@mail.sysu.edu.cn}
\credit{Supervision}

\author[1]{Hanjiang Lai}
\cormark[1]
\ead{laihanj3@mail.sysu.edu.cn}
\credit{Writing – review \& editing, Supervision}

\affiliation[1]{department={School of Computer and Engineering,},
                organization={Sun Yat-sen University},
                city={Guangzhou},
                postcode={510000}, 
                country={China}}

\affiliation[2]{department={School of Artificial Intelligence,},
                organization={Sun Yat-sen University},
                city={Zhuhai},
                postcode={519000}, 
                country={China}}

\cortext[cor1]{Corresponding author}

\begin{abstract}
Out-of-distribution (OOD) detection is paramount to ensuring the reliability and robustness of learning models in real-world applications. Existing post-hoc OOD detection methods detect OOD samples by leveraging their features and logits information without retraining. However, they often overlook the inherent correlation between features and logits, which is crucial for effective OOD detection. To address this limitation, we propose Global-Aware Feature Decoupling with Confidence Calibration (GAFD-CC). GAFD-CC aims to refine decision boundaries and increase discriminative performance. Firstly, it performs global-aware feature decoupling guided by classification weights. This involves aligning features with the direction of global classification weights to decouple them. From this, GAFD-CC extracts two types of critical information: positively correlated features that promote in-distribution (ID)/OOD boundary refinement and negatively correlated features that suppress false positives and tighten these boundaries. Secondly, it adaptively fuses these decoupled features with multi-scale logit-based confidence for comprehensive and robust OOD detection. Extensive experiments on large-scale benchmarks demonstrate GAFD-CC's competitive performance and strong generalization ability compared to those of state-of-the-art methods.
\end{abstract}




\maketitle

\section{Introduction}

Machine learning models, typically trained in closed environments, inevitably encounter out-of-distribution (OOD) data when deployed in real-world scenarios (\cite{Open_World2006}). In such cases, these models can produce confidently incorrect predictions (\cite{7298640}). This issue poses significant risks to safety-critical domains, such as autonomous driving (\cite{5548123}) and medical analysis (\cite{schlegl2017unsupervisedanomalydetectiongenerative}), necessitating effective OOD detection methods to enhance model safety. Currently, the predominant approach to OOD detection involves designing a post-hoc score function (\cite{hendrycks2018baselinedetectingmisclassifiedoutofdistribution, liang2020enhancingreliabilityoutofdistributionimage, wang2022vimoutofdistributionvirtuallogitmatching}) that computes a confidence score to identify OOD samples without requiring model retraining.

Existing post-hoc OOD detection methods generally fall into logit-based (\cite{hendrycks2018baselinedetectingmisclassifiedoutofdistribution, liu2021energybasedoutofdistributiondetection}) and feature-based (\cite{sun2021reactoutofdistributiondetectionrectified, sun2022diceleveragingsparsificationoutofdistribution}) categories (\cite{ling2025cadrefrobustoutofdistributiondetection}). Logit-based methods are simple and efficient, leveraging the classifier's output confidence. Still, they can be fragile when the model exhibits overconfidence or when In-Distribution(ID) and OOD samples have similar logits. In contrast, feature-based methods capture richer semantic information from intermediate layers but often struggle to find an optimal distance metric or calibration strategy to separate ID and OOD samples effectively. To overcome the inherent limitations of these individual approaches and fully utilize their complementary strengths, hybrid methods (\cite{liang2020enhancingreliabilityoutofdistributionimage, wang2022vimoutofdistributionvirtuallogitmatching, ling2025cadrefrobustoutofdistributiondetection}) have emerged, synthesizing various sources of information such as logits, features, and gradients. For example, ODIN (\cite{liang2020enhancingreliabilityoutofdistributionimage}) improves logit discriminability through gradient-based perturbation. ViM (\cite{wang2022vimoutofdistributionvirtuallogitmatching}) generates virtual OOD logits from feature residuals. OptFS (\cite{zhao2024optimalfeatureshapingmethodsoutofdistribution}) influences logits through feature shaping. CADRef (\cite{ling2025cadrefrobustoutofdistributiondetection}) decouples features based on max-logit weights. However, many existing hybrid methods, including the aforementioned ones, still employ static or limited fusion strategies. They typically rely on restricted directional projections or fixed rules that lack adaptive calibration and have yet to fully exploit the inherent global correlation between the entire feature vector and the complete set of classification weights, which may constrain their performance. This gap raises a critical question: can we better exploit this global correlation for more robust OOD detection?

To this end, we propose Global-Aware Feature Decoupling with Confidence Calibration (GAFD-CC). GAFD-CC enhances OOD detection performance through a two-step process. Firstly, it performs global-aware feature decoupling guided by classification weights. This aligns features with the direction of global classification weights, extracting two types of critical information: positively correlated features that enhance the separation and clarity of ID/OOD boundaries, and negatively correlated features that suppress false positives and robustify these boundaries. Secondly, it adaptively fuses these decoupled features with multi-scale logit-based confidence measures, forming OOD detection scores that are dynamically weighted to boost performance for comprehensive and robust OOD detection significantly.

In summary, the main contributions of this paper are as follows:

\begin{itemize}
\item We propose GAFD-CC, a novel OOD detection framework that effectively integrates multi-scale logits information with features by introducing Global-Aware Feature Decoupling to capture their inherent global interplay.

\item We demonstrate that GAFD-CC leverages the directional deviations of features with respect to classification weights to derive novel discriminative scores, which intrinsically capture and enhance the separation between in-distribution and OOD samples by significantly reducing score distribution overlap.

\item Through extensive experimental evaluations on diverse benchmarks, including challenging hard OOD datasets and various CNN- and Transformer-based architectures, we demonstrate that our method achieves competitive performance against state-of-the-art baselines and exhibits strong robustness.
\end{itemize}

\section{Related Work}

\subsection{OOD data exposure}

One major research direction involves utilizing auxiliary Out-of-Distribution (OOD) data during training, a technique pioneered by Outlier Exposure (\cite{DBLP:journals/corr/abs-1812-04606}). The core idea is to train the model to produce less confident or uniform predictions on these known outlier samples, leading to strategies like feature space regularization (\cite{dhamija2018reducingnetworkagnostophobia, ming2022exploit, huang2021mosscalingoutofdistributiondetection}) and synthetic data generation (\cite{lee2018simpleunifiedframeworkdetecting, yang2021semanticallycoherentoutofdistributiondetection}). Applying OE in the context of federated learning (\cite{jeong2025out}) is an extension of this direction. However, the effectiveness of these methods is fundamentally tethered to the representativeness of the auxiliary data. This dependency can induce bias toward seen OOD types, hinder generalization under domain discrepancies due to feature space misalignment, and incur prohibitive computational costs from complex training frameworks. These unresolved issues motivate the development of data-agnostic, efficient, and domain-robust post-hoc solutions, which are the focus of our work.

\subsection{Post-hoc OOD score} 

Post-hoc OOD detection is a methodology that identifies OOD samples by computing a confidence score without requiring model retraining. Post-hoc OOD detection methods primarily evolve along three complementary yet limited paradigms: logit-based methods, feature-based methods, and hybrid methods (\cite{ling2025cadrefrobustoutofdistributiondetection}). 

Logit-based methods primarily use the model's logit outputs to construct confidence scores (\cite{wei2022mitigatingneuralnetworkoverconfidence, 10204972}). Common techniques include compressing the logit vector into a single scalar, such as the maximum softmax probability (MSP) (\cite{hendrycks2018baselinedetectingmisclassifiedoutofdistribution}) or energy over logits (\cite{liu2021energybasedoutofdistributiondetection}). Despite their simplicity, a key limitation is their vulnerability to overconfidence bias, which often leads to failures when OOD samples yield spuriously high logits.

Feature-based methods analyze deviations within the feature space, typically from intermediate layers, to detect OOD samples (\cite{ahn2023lineoutofdistributiondetectionleveraging}). Many adopt a “project-then-measure” philosophy. ReAct (\cite{sun2021reactoutofdistributiondetectionrectified}) truncates extreme activations to normalize the feature distributions, making in-distribution features more concentrated while pushing OOD features further away. DICE (\cite{sun2022diceleveragingsparsificationoutofdistribution}) sparsifies important weights by removing or reducing the influence of less important features to enhance the signal-to-noise ratio in feature representations, making the core discriminative features more prominent for OOD detection. 

Hybrid methods aim to enhance performance by combining scores from multiple sources. ODIN (\cite{liang2020enhancingreliabilityoutofdistributionimage}) enhances logit discriminability by applying a subtle perturbation to input data, derived from gradient information with respect to the maximum logit. GradVec (\cite{carvalho2025towards}) leverages the gradient space by proposing a family of methods that use the gradient vector as a sparse input representation for OOD detection, arguing that the gradient provides more informative knowledge for distinguishing unknown samples. ViM (\cite{wang2022vimoutofdistributionvirtuallogitmatching}) leverages both feature and logit spaces by first decomposing the feature space into class-relevant and class-agnostic subspaces. It then projects a sample's feature residuals onto the class-agnostic subspace and synthesizes a ``virtual OOD logit” based on the magnitude of this projection, ultimately combining it with standard classification logits for a stronger OOD score. CADRef (\cite{ling2025cadrefrobustoutofdistributiondetection}) integrates feature and logit information by decoupling class-aware relative features into positive and negative components aligned with the maximum-logit weight, and subsequently rescaling each component based on confidence scores. However, many existing hybrid methods, including those discussed above, still exhibit limitations in their information fusion strategies, failing to fully exploit or adaptively leverage multi-source information. This leaves considerable room for the development of more robust and adaptive OOD detection approaches.

\begin{figure*}[t]
\centering
\includegraphics[width=0.8\textwidth]{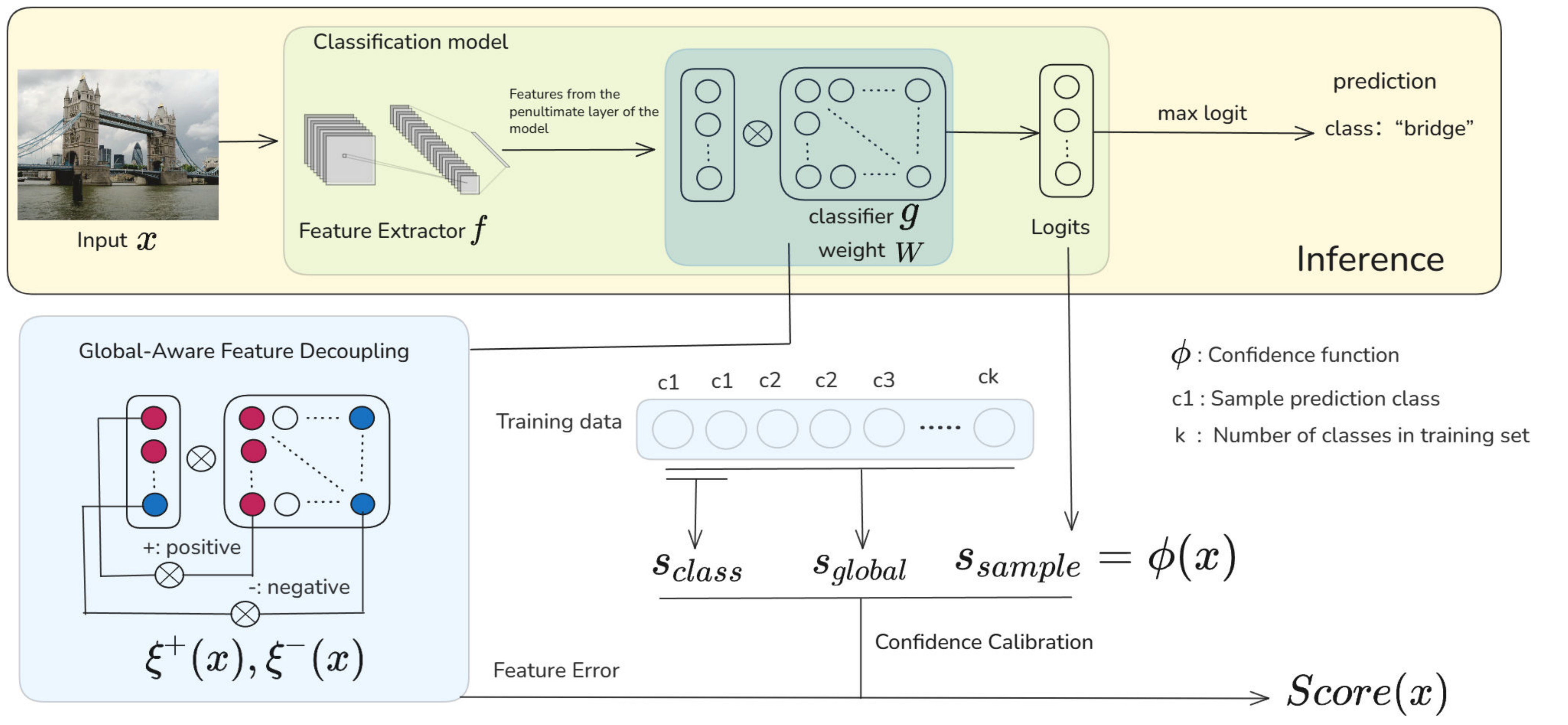} 
\caption{GAFD-CC Pipeline. Our proposed GAFD-CC framework involves two main steps: Global-Aware Feature Decoupling, which decouples features into positively/negatively correlated components to compute feature error scores ($\xi ^{+}(x), \xi ^{-}(x)$); and Confidence Calibration, which adaptively fuses logit-based scores ($s_{sample}, s_{class}, s_{global}$) at different scales with feature error scores to yield the final OOD detection score ($Score(x)$).}
\label{fig:pipeline}
\end{figure*}

\section{Preliminaries}

\subsection{OOD detection problem formulation}

Consider a classification model $F$ trained on an ID dataset $D_{train}$ , where samples $x \in X$ are mapped to an output space $Y=\{1,…,k \}$, with $k$ representing the number of ID classes. The neural network $F$ can be decomposed into a feature extractor $f$ and a classifier $g$, such that $F = f\circ g$. Here, $f(x)$ denotes the feature vector extracted from the input $x$, and the classifier $g$ (typically a linear layer with weight matrix $W$ and bias vector $b$) computes logit outputs based on these features.

The goal of OOD detection is to identify samples that do not originate from the training data distribution $D_{train}$. These samples are referred to as OOD samples, in contrast to ID samples, which are drawn from $D_{train}$. This task is typically framed as a binary classification problem (ID vs. OOD) based on a confidence scoring function $SCORE(\cdot)$. For a given input $x$, the scoring function assigns a scalar value, such that:
\begin{equation}
    \begin{cases}
    x \in ID, \quad SCORE(x)>\tau, \\
    x \in OOD, \quad SCORE(x) \le \tau, 
    \end{cases}
\end{equation}
where $\tau$ is a predetermined threshold. A sample is classified as an OOD sample if its computed score falls below $\tau$; otherwise, it is classified as an ID sample.

\subsection{CADRef and further considerations} 

The Class-Aware Decoupled Relative Feature-based method (CADRef) (\cite{ling2025cadrefrobustoutofdistributiondetection}) is a recent approach that improves OOD detection by decoupling features. CADRef operates by transforming raw features into class-aware relative features by subtracting the mean feature of the predicted class. It then proceeds to decouple these relative features into positive and negative correlation components. This decoupling is based on whether the sign of each relative feature dimension aligns with the corresponding classification weight sign of the maximum logit. Subsequently, it rescales each decoupled component based on confidence scores. Through this systematic approach, CADRef effectively enhances OOD sample detection performance by leveraging both feature and logit information. 

However, CADRef raises two considerations for robust OOD detection. Neural networks can produce arbitrarily high softmax confidence for inputs far from the training distribution (\cite{7298640}), so logit-magnitude or energy-based scores (\cite{liu2021energybasedoutofdistributiondetection}) are usually preferred. Therefore, CADRef anchors feature decoupling to the weight vector of the class with the maximum logit, projecting the feature space onto that single axis and discarding all other class-specific directions. This one-dimensional projection intrinsically discards information and amplifies vulnerability: when an adversarial or out-of-distribution sample produces a spurious maximum, every subsequent operation follows the wrong axis, cascading into a confident misclassification. Moreover, a fixed global scaling factor prevents the anomaly score from adapting to local statistics, further compounding rigidity. Consequently, when the largest logit is large but the remaining logits are nearly uniform, CADRef still treats the isolated peak as in-distribution evidence, turning model overconfidence into missed anomalies.

\section{Method}

In this section, we detail our proposed Global-Aware Feature Decoupling with Confidence Calibration (GAFD-CC). We first introduce Global-Aware Feature Decoupling (GAFD), which globally considers the interplay between features and logits information to enhance the separability of in-distribution (ID) and OOD samples. Given that for classification models, logits information is obtained by multiplying features with the fully connected classification weights of the last layer, GAFD captures this correlated interplay by analyzing the directional consistency between features and the classification weight vectors. The term ``Global'' refers to the fact that our method considers the classification weights of all classes to guide the feature decoupling process.

Subsequently, we propose a robust Confidence Calibration (CC) approach that judiciously fuses the decoupled features obtained from GAFD with multi-scale logit information as a confidence measure. This approach, known as GAFD-CC, achieves improved OOD detection performance by effectively leveraging both feature and confidence signals. The pipeline of our proposed method is illustrated in Figure~\ref{fig:pipeline}. Below, we will progressively detail our implementation steps.

\subsection{Global-aware feature decoupling} 

For input sample $x$, we first extract the deep feature vector $f(x)\in R^d$ using a Classification model. To better perform OOD detection, we utilize the deviation between sample features and the predicted class of the sample as feature space information (\cite{ling2025cadrefrobustoutofdistributiondetection}) for subsequent calculations. Following its paradigm, we compute the feature deviation from the predicted class c center:
\begin{equation}
\Delta f(x) = f(x) - \mu_{c},
\end{equation}
where $c$ is the class predicted by the model for the input sample $x$, and $\mu_{c}$ is the pre-computed mean feature vector of all training samples belonging to class $c$.

We then perform decoupling by quantifying the relationship between the classifier weight matrix and the feature deviation. The weight matrix is $W \in R^{K \times d}$, and the bias vector is $b \in R^{K}$, where $K$ is the number of classes. Then the logits vector $s \in R^{K}$ is computed as:
\begin{equation}
    s = W \Delta f(x)+b.
\end{equation}
The $j$-th logit (for class $j$) is :
\begin{equation}
    s_{j}=\sum^{d}_{i=1}W_{j, i} \cdot \Delta f_{i}+b_{j}.
\end{equation}
For assessing the global impact of each feature component $\Delta f_{i}$ (for $i=1,...,d$) on the entire logits vector s,
\begin{equation}
    s = [s_{1:K}]^{\mathstrut T} = \Biggl[ \Bigl( \sum_{i=1} w_{j,i} \cdot \Delta f_i + b_j \Bigr)_{\!\!j=1:K} \Biggr]^{\mathstrut T}.
\end{equation}
The partial derivative of the eigencomponent $\Delta f_{i}$ to $s_{sum}$ is:
\begin{equation}
    \frac{\partial s}{\partial \Delta f_{i}}=\sum^{K}_{j=1}W_{j,i}.
\end{equation}
Let $w_{i}=\sum^{K}_{j=1}W_{j,i}$ be the sum of the weights in the $i$-th column of $W$, which represents the sensitivity of the logits vector ($s$) to changes in the i-th feature component ($\Delta f_{i}$). The bias is constant and does not depend on $\Delta f$. Therefore, the sign of $w_{i}\cdot \Delta f_{i}$ determines the effect of $\Delta f_{i}$ on the logits vector: a positive value ($w_{i}\cdot \Delta f_{i}>0$) indicates that $\Delta f_{i}$ will increase the $s$, while a negative value ($w_{i}\cdot \Delta f_{i}<0$)  indicates it will decrease the logits vector. Hence we classify each feature component $\Delta f_{i}$ as: 
\begin{equation}
\begin{cases}
\Delta f^{+}(x)_{i} \quad \text{if } w_{i}\cdot \Delta f_{i}>0 ,\\
\Delta f^{-}(x)_{i} \quad \text{if } w_{i}\cdot \Delta f_{i}<0 .
\end{cases}
\end{equation}
$\Delta f^{+}(x)$ and $\Delta f^{-}(x)$ denote the positively correlated feature component and the negatively correlated feature component.
This formulation reflects the observation that aligned signs between weights and feature deviations contribute positively to logits, while sign mismatches cause adversarial effects that reduce logit values and impair OOD detection.
Direction-sensitive scores are subsequently computed:
\begin{equation}
\begin{aligned}
\xi ^{+}(x) = \frac{||\Delta f^{+}(x)||_{1}}{||f(x)||_{1}}, 
\xi ^{-}(x) = \frac{||\Delta f^{-}(x)||_{1}}{||f(x)||_{1}}.
\end{aligned}
\end{equation}
The L1 norm enhances sensitivity to directional deviations, while normalization by $||f(x)||_{1}$ scales the scores based on feature magnitude.

\begin{figure}
\centering
    \begin{subfigure}[b]{0.45\columnwidth}
        \includegraphics[width=\textwidth]{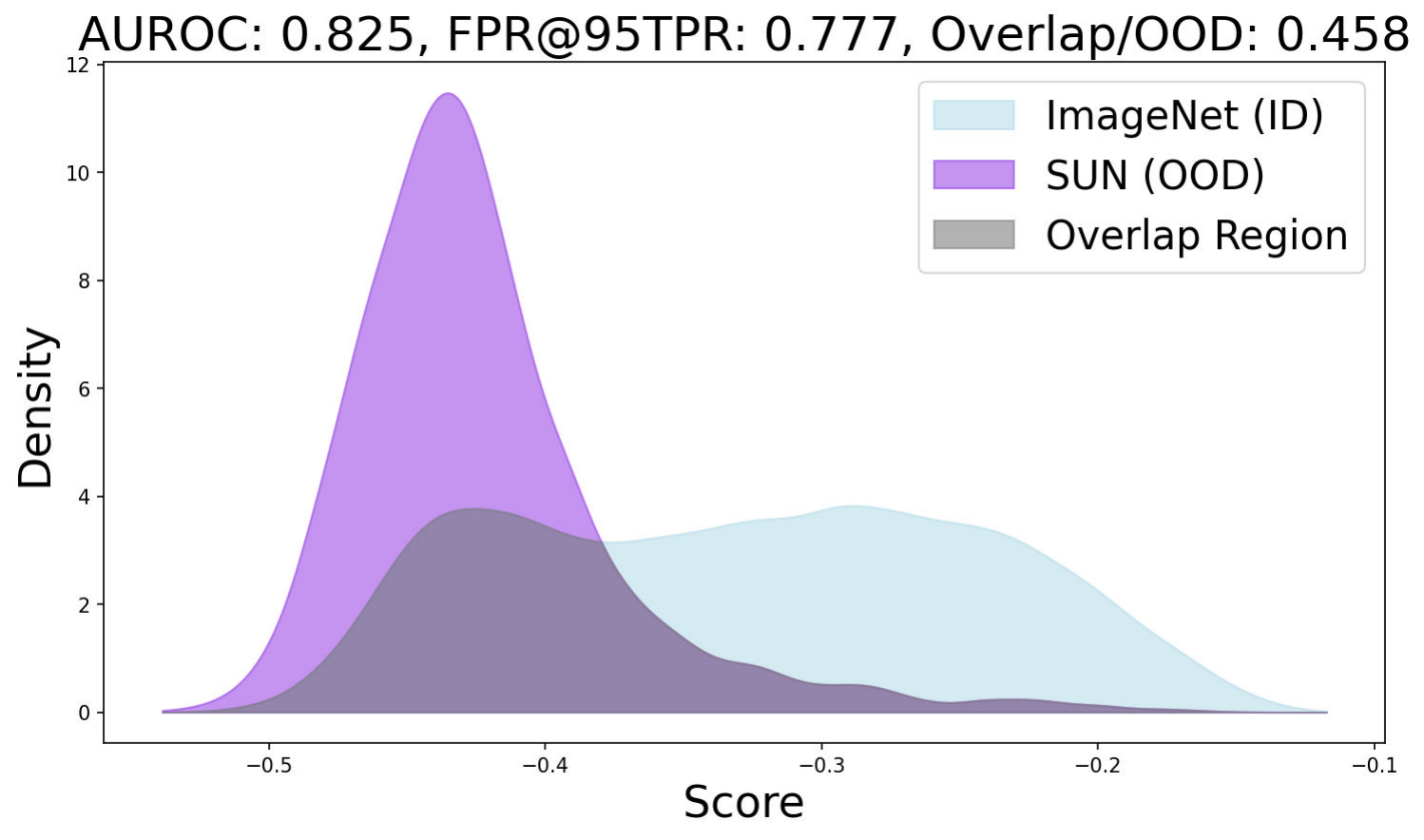}
        \caption{}
        \label{fig:a}
    \end{subfigure}
    \hfill
    \begin{subfigure}[b]{0.45\columnwidth}
        \includegraphics[width=\textwidth]{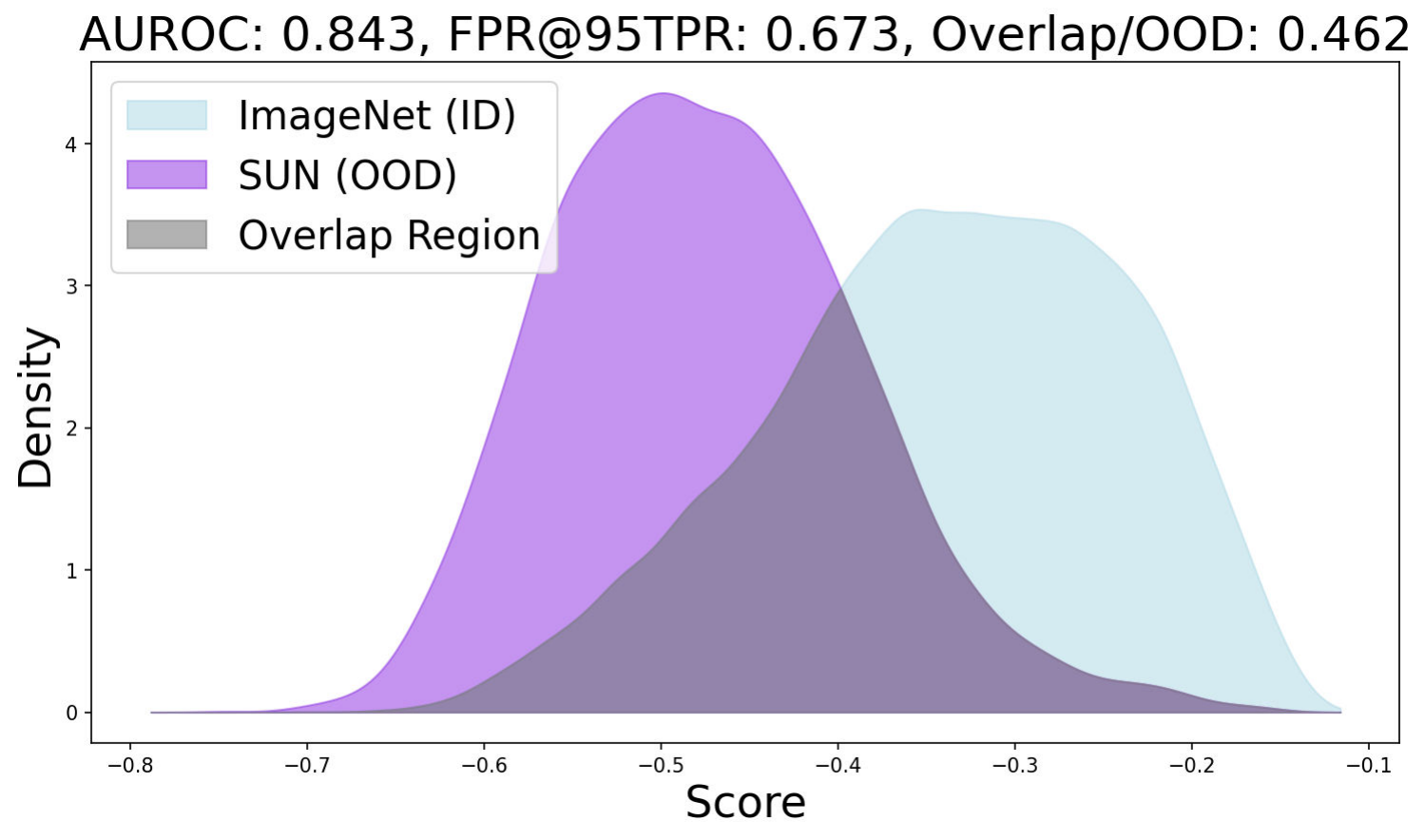}
        \caption{}
        \label{fig:b}
    \end{subfigure}

    \begin{subfigure}[b]{0.45\columnwidth}
        \includegraphics[width=\textwidth]{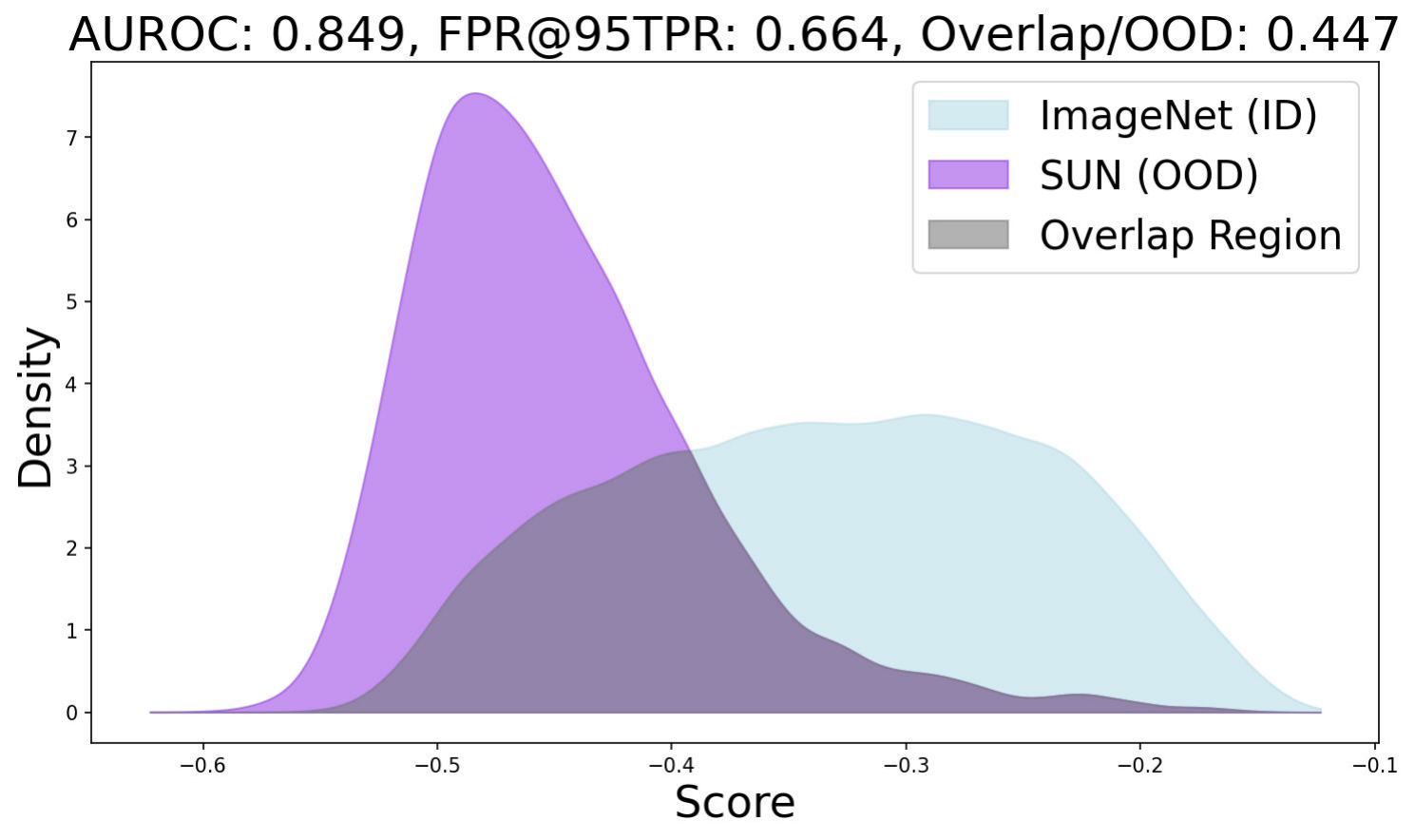}
        \caption{}
        \label{fig:c}
    \end{subfigure}
    \hfill
    \begin{subfigure}[b]{0.45\columnwidth}
        \includegraphics[width=\textwidth]{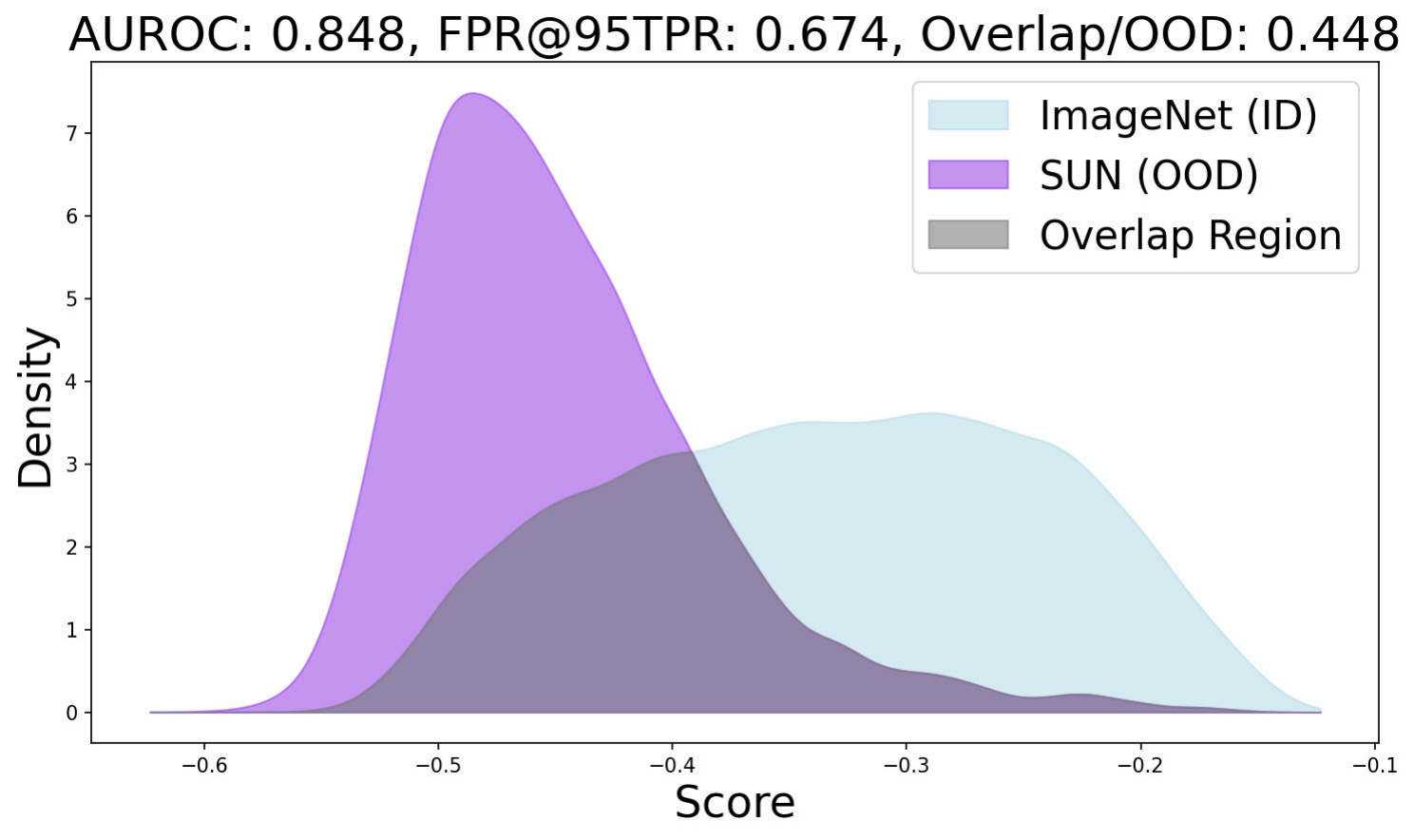}
        \caption{}
        \label{fig:d}
    \end{subfigure}

\caption{Comparison of score distributions for ID (ImageNet-1K, blue) and OOD (SUN, purple) samples. (a, b) show the positive and negative feature scores from CADRef. (c, d) show the scores from our method. Our method demonstrates better separation (less overlap) between ID and OOD distributions.}
\label{fig:compare}
\end{figure}

\subsubsection{Discussion} 

While inspired by CADRef's use of positive and negative feature components (\cite{ling2025cadrefrobustoutofdistributiondetection}), our Global-Aware Feature Decoupling (GAFD)fundamentally differentiates itself by analyzing the interplay between features and the entire set of classification weights ($W$), instead of relying on a single maximum-logit direction for guidance. A direct comparison of the feature distributions in Figure~(\ref{fig:a},\ref{fig:b}) with Figure~(\ref{fig:c},\ref{fig:d}) reveals that our GAFD method achieves a significantly superior separation between ID and OOD samples compared to the CADRef approach. This clear reduction in the overlapping feature space directly translates to a notable enhancement in discriminability. Consequently, our results validate that the proposed GAFD effectively minimizes this critical overlap, demonstrating a distinct advantage over the CADRef feature decoupling mechanism.

\begin{table*}[t]
\centering

\resizebox{\textwidth}{!}{%
\begin{tabular}{lcccccccccccccccc}
\toprule
\multirow{2}{*}{Methods} & \multirow{2}{*}{Source} & \multicolumn{2}{c}{iNaturalist} & \multicolumn{2}{c}{SUN} & \multicolumn{2}{c}{Places} & \multicolumn{2}{c}{Textures} & \multicolumn{2}{c}{OpenImage-O} & \multicolumn{2}{c}{ImageNet-O} & \multicolumn{2}{c}{Average} \\
\cmidrule(lr){3-4}\cmidrule(lr){5-6}\cmidrule(lr){7-8}\cmidrule(lr){9-10}\cmidrule(lr){11-12}\cmidrule(lr){13-14}\cmidrule(lr){15-16}
& & AUROC$\uparrow$ & FPR95$\downarrow$ & AUROC$\uparrow$ & FPR95$\downarrow$ & AUROC$\uparrow$ & FPR95$\downarrow$ & AUROC$\uparrow$ & FPR95$\downarrow$ & AUROC$\uparrow$ & FPR95$\downarrow$ & AUROC$\uparrow$ & FPR95$\downarrow$ & AUROC$\uparrow$ & FPR95$\downarrow$ \\
\midrule
MSP  & logit & 88.39 & 52.84 & 81.64 & 69.11 & 80.53 & 72.07 & 80.43 & 66.28 & 84.85 & 64.06 & 28.62 & 100.00 & 74.08 & 70.73 \\
MaxLogit & logit & 91.15 & 50.80 & 86.44 & 60.41 & 84.03 & 66.06 & 86.39 & 54.95 & 89.13 & 57.91 & 40.74 & 100.00 & 79.65 & 65.02 \\
ODIN & grad+logit & 87.01 & 52.40 & 86.57 & 53.63 & 85.30 & 58.63 & 86.51 & 46.01 & 86.65 & 52.82 & 43.65 & 98.60 & 79.28 & 60.35 \\
Energy & logit & 90.62 & 53.77 & 86.52 & 58.84 & 83.96 & 66.03 & 86.72 & 52.48 & 89.02 & 57.70 & 41.79 & 100.00 & 79.78 & 64.80 \\
ReAct & feat & 96.42 & 19.30 & \textbf{94.42} & \textbf{24.18} & \textbf{91.93} & \textbf{33.64} & 90.46 & 45.83 & 90.56 & 43.77 & 52.46 & 98.05 & 86.04 & 44.13 \\
ViM & feat+logit & 87.43 & 71.84 & 81.07 & 81.82 & 78.39 & 83.14 & 96.83 & 14.88 & 89.31 & 58.70 & \underline{70.78} & \textbf{84.90} & 83.97 & 65.88 \\
DICE & feat & 94.51 & 26.63 & 90.92 & \underline{36.47} & 87.65 & \underline{47.97} & 90.44 & 32.61 & 88.57 & 45.72 & 42.78 & 98.00 & 82.48 & 47.90 \\
GEN & logit & 92.44 & 45.76 & 85.52 & 65.54 & 83.46 & 69.24 & 85.42 & 60.32 & 89.22 & 60.96 & 43.60 & 100.00 & 79.94 & 66.97 \\
OptFS & feat+logit & \underline{96.83} & \underline{17.03} & \underline{93.05} & \underline{35.70} & \underline{90.32} & \underline{45.24} & 95.68 & 23.49 & 92.69 & 38.02 & 59.71 & 97.35 & 88.05 & 42.80 \\
CADRef & feat+logit & \underline{96.90} & \underline{16.07} & 91.26 & 39.24 & 87.81 & 51.13 & \underline{97.15} & \underline{12.64} & \underline{93.94} & \underline{32.68} & 68.39 & 92.35 & \underline{89.24} & \underline{40.69} \\
\midrule
\textbf{GAFD-C} & feat+logit & \textbf{96.99} & \textbf{15.56} & 91.15 & 39.83 & 87.57 & 51.85 & \underline{97.34} & \underline{11.91} & \textbf{93.99} & \textbf{32.52} & \underline{69.92} & \underline{89.90} & \underline{89.49} & \underline{40.26} \\
\textbf{GAFD-CC} & feat+logit & 96.43 & 18.44 & \underline{92.00} & 37.16 & \underline{88.75} & 48.63 & \textbf{97.63} & \textbf{10.98} & \underline{92.95} & \underline{36.04} & \textbf{71.84} & \underline{86.70} & \textbf{89.93} & \textbf{39.66} \\
\bottomrule
\end{tabular}
}
\caption{Results of OOD detection on ImageNet-1K benchmark with ResNet-50. $\uparrow$ indicates that higher values are better, while $\downarrow$ indicates that lower values are better. All values are percentages, with the best results being \textbf{highlighted} , the 2nd and 3rd ones being \underline{underlined}, respectively.The source usage definition is described in section \textbf{Baseline methods.}}
\label{tab:ood_detection_resnet}
\end{table*}

\subsection{Logit-based confidence calibration} 

To optimally leverage the decoupled components, we introduce a confidence calibration scheme that treats positive and negative deviations differently, reflecting their distinct roles in OOD detection. Recognizing that relying solely on sample features can limit OOD detection effectiveness (\cite{wang2022vimoutofdistributionvirtuallogitmatching}) and that integrating logits information is crucial, we adopt a multi-source hybrid approach. Specifically, we explore the interplay between features and logits information, utilizing established logit-based scores (Energy (\cite{liu2021energybasedoutofdistributiondetection})) to enhance our proposed GAFD-CC framework.

We incorporate logit-based confidence at multiple scales while avoiding hyperparameter sensitivity through a hierarchical confidence framework: the sample confidence $s_{sample}=\phi(x)$, the class-mean confidence $s_{class} = E_{x \in D_{c}}[\phi(x)]$, and the global-mean confidence $s_{global} = E_{x \in D_{train}}[\phi(x)]$, where the confidence function $\phi(.)=-log\sum (exp(s/T))$ is a logit score function from Energy (\cite{liu2021energybasedoutofdistributiondetection}), $D_{c}$ denotes the set of ID training samples belonging to the predicted class $c$ of sample $x$, and $D_{train}$ represents the entire set of ID training samples. For each test sample $x$, the sample confidence $s_{sample}$ is the current sample's energy score. It captures the model's detection confidence for that sample. Class-mean confidence $s_{class}$, which is derived from the average logit values of training samples belonging to the predicted class, characterizes the prediction stability at the class level. Global-mean confidence $s_{global}$, which is based on the average logit values of all training samples, establishes a baseline for the model's overall logit information. Both class-mean confidence and global-mean confidence are pre-computed during training.

Since samples with high positively correlated feature scores are considered ``over-matched” classification deceptive samples, studies (\cite{ling2025cadrefrobustoutofdistributiondetection}) have shown that positive features effectively distinguish between ID and OOD samples at lower logit values. In these cases, we utilize sample confidence and class-mean confidence to normalize these scores, thereby enhancing the discriminability between ID and OOD samples. Conversely, negatively correlated features are primarily employed to capture typical OOD samples that deviate from classification weights. For these, we normalize the scores using global-mean confidence and class-mean confidence, which, by considering highly correlated weight directions and leveraging inter-class relationships, enhances the perception of unknown class distributions and improves the discriminability of these types of samples.

The final score formula of our GAFD-CC is as follows:
\begin{equation}
\begin{aligned}
Score(x) = \frac{\xi^{+}(x)}{s_{sample} + s_{class}} + 
\frac{\xi^{-}(x)}{s_{global}+s_{class}}.
\end{aligned}
\end{equation}
$Score(x)$ is an adaptive fusion of three confidence measures. While the two feature-based terms, $\xi^{+}(x)$  and $\xi^{-}(x)$, have distinct normalization terms, they are weighted equally (1:1) with respect to the initial combination coefficient in the final score derivation. The ablation study later investigates the optimal choice of these coefficients.

\subsection{Inference} 

During inference, GAFD-CC efficiently computes a $Score(x)$ for sample $x$. The components of this score, specifically the feature scores ($\xi^{+}(x),\xi^{-}(x)$) and the sample logit confidence ($s_{sample}$), are derived from the pre-trained classification model's intermediate features and final logit outputs. Crucially, additional statistics such as each class's mean feature ($\mu_{c}$), global mean logit confidence ($s_{global}$), and each class's mean logit confidence ($s_{class}$) are pre-computed during the training phase. Based on this $Score(x)$, a sample is classified as an ID sample if it is above a pre-set threshold $\tau$; otherwise, it is deemed OOD. This streamlined process enables highly efficient OOD detection.

\section{Experiments}

\subsection{Experimental setting}

\subsubsection{Dataset}

For OOD detection benchmark, we use ImageNet-1K (\cite{2009ImageNet}) as ID data and six commonly used OOD datasets: iNaturalist (\cite{2018The}), SUN (\cite{2010SUN}), Places (\cite{2018Places}), Textures (\cite{cimpoi2013describingtextureswild}), OpenImage-O (\cite{wang2022vimoutofdistributionvirtuallogitmatching}) and ImageNet-O (\cite{hendrycks2021naturaladversarialexamples}). Covering natural species, scene contexts, material textures, and adversarially curated images, these benchmarks span diverse visual domains and semantic categories, thereby providing a comprehensive test-bed for evaluating the generalizability of OOD detection methods.

\subsubsection{Baseline methods}

We compare the proposed GAFD-CC method against ten existing baseline approaches. The comparative methods encompass three distinct information sources: logit-based methods, feature-based methods, and hybrid methods. The logit-based methods include MSP (\cite{hendrycks2018baselinedetectingmisclassifiedoutofdistribution}), MaxLogit (\cite{hendrycks2022scalingoutofdistributiondetectionrealworld}), Energy (\cite{liu2021energybasedoutofdistributiondetection}), and GEN (\cite{Liu2023GENPT}). The feature-based methods include ReAct (\cite{sun2021reactoutofdistributiondetectionrectified}) and DICE (\cite{sun2022diceleveragingsparsificationoutofdistribution}). The hybrid methods include: ODIN (\cite{liang2020enhancingreliabilityoutofdistributionimage}) (grad+logit), which employs gradient information to enhance logit score separation; OptFS (\cite{zhao2024optimalfeatureshapingmethodsoutofdistribution}) (feat+logit), which adapts features layer-wise to promote logit score separation;  ViM (\cite{wang2022vimoutofdistributionvirtuallogitmatching}) (feat+logit), which constructs virtual OOD logits using class-agnostic feature residuals and CADRef (\cite{ling2025cadrefrobustoutofdistributiondetection}) (feat+logit), which refines feature representations through component-aware decomposition and distribution alignment.

\subsubsection{Evaluation metrics}

We adopt AUROC and FPR95 as the evaluation metrics for OOD detection. AUROC denotes the area under the ROC curve, which plots the True Positive Rate against the False Positive Rate at various thresholds. FPR95 is the false-positive rate among OOD samples when 95\% of ID samples are correctly identified. A higher AUROC and a lower FPR95 indicate superior OOD detection performance.

\subsection{Quantitative evaluations}

\subsubsection{The OOD detection benchmark}

Table~\ref{tab:ood_detection_resnet} summarizes the out-of-distribution detection performance of each method on various OOD datasets and the average performance over the six OOD sets for the ImageNet-1K benchmark. GAFD-C represents the result obtained without using class-mean confidence for adjustment, the detailed discussion of this part is in Hyperparameter discussion.

\textbf{Comparison with Logit-based Methods.} GAFD-CC consistently outperforms all logit-based methods. Specifically, GAFD-CC achieves an Average AUROC of 89.93\%, which is higher than the best logit-based method, GEN (79.94\%). Furthermore, GAFD-CC demonstrates a lower Average FPR95 of 39.66\% compared to GEN's 66.97\%, highlighting its superior ability to distinguish between in-distribution and out-of-distribution samples with a reduced false positive rate.

\textbf{Comparison with Feature-based Methods.} Feature-based methods like ReAct and DICE aim to capture richer semantic information, but they exhibit notable performance variability. For instance, ReAct achieves an average AUROC of 86.04\% and FPR95 of 44.13\%, while DICE's average AUROC is 82.48\% with an FPR95 as high as 47.90\%. In contrast, our GAFD-CC method effectively integrates deep features from intermediate layers with the final logit information. This comprehensive approach significantly enhances OOD detection performance and robustness, allowing GAFD-CC to surpass both ReAct and DICE in key metrics.

\textbf{Comparison with Hybrid Methods.} GAFD-CC demonstrates robust performance compared to other hybrid approaches. It reduces the average FPR95 to 39.66\%, outperforming ODIN's 60.35\% and OptFS's 42.80\%. While ViM achieves exceptional OOD detection (AUROC 83.97\%, FPR95 65.88\%), its mechanism introduces additional design complexity. Our GAFD-CC provides a strong balance of performance without relying on such specific, complex designs. Furthermore, GAFD-CC slightly surpasses CADRef's average AUROC (89.93\% vs 89.24\%) and significantly lowers its average FPR95 (39.66\% vs 40.69\%), confirming the effectiveness of GAFD-CC's hybrid strategy for more precise OOD discrimination and reliable real-world performance.

\begin{table}[t]
\centering
\resizebox{\columnwidth}{!}{
\begin{tabular}{lcccccc}
\toprule
\multirow{2}{*}{Methods}  & \multicolumn{2}{c}{SSB-hard} & \multicolumn{2}{c}{NINCO} \\
\cmidrule(lr){2-3}\cmidrule(lr){4-5}
& AUROC$\uparrow$ & FPR95$\downarrow$  & AUROC$\uparrow$ & FPR95$\downarrow$ \\
\midrule

MaxLogit   & 65.04 & 81.34 & 72.31 & 71.84 \\
ODIN    & 58.81 & 92.73 & 61.05 & 91.20 \\
ViM   & 72.09 & 84.76 & 82.32 & 72.93 \\
DICE  & 50.17 & 92.18 & 38.54 & 91.03 \\
GEN  & 72.49 & 80.64 & 82.78 & \underline{64.73} \\
OptFS & 68.95 & 86.03 & 80.60 & 74.91 \\
CADRef   & \underline{74.70} & \underline{79.67} & \underline{85.25} & 65.41 \\
\midrule
\textbf{GAFD-CC(Ours)}  & \textbf{74.98} & \textbf{78.28} & \textbf{85.52} & \textbf{64.14} \\
\bottomrule
\end{tabular}
}
\caption{Results of two Near-OOD datasets on ImageNet-1K benchmark. All values are percentages, with the best results being \textbf{highlighted} , the second best being \underline{underlined}, respectively.}
\label{tab:hard_ood}
\end{table}

\subsubsection{Robustness discussion}

We evaluated the robustness of GAFD-CC using SSB-hard (\cite{vaze2022opensetrecognitiongoodclosedset}) and NiNCO (\cite{bitterwolf2023outfixingimagenetoutofdistribution}). These benchmarks represent a stringent test of robustness as they are near-OOD datasets, specifically designed to be visually and semantically confusable with the ImageNet-1K in-distribution (ID) data. Distinguishing these samples requires a model to be highly sensitive to subtle deviations from the ID manifold, rather than just identifying obvious anomalies.As demonstrated in Table~\ref{tab:hard_ood}, GAFD-CC exhibits exceptional robustness by establishing new state-of-the-art performance on these challenging datasets. Our method decisively outperforms all seven representative baselines: 74.98\% AUROC / 78.28\% FPR95 on SSB-hard and 85.52\% AUROC / 64.14\% FPR95 on NINCO. This performance surpasses the strong second-best method, CADRef (74.70\%/85.25\% AUROC), and highlights a significant performance gap over other established methods like ViM (72.09\%/82.32\% AUROC). This leading performance underscores our method's robust capability. Notably, GAFD-CC achieves this result without the high algorithmic complexity of competing methods. ViM, for instance, requires an intricate process of PCA, subspace projection, and virtual logit generation. In contrast, GAFD-CC's robustness stems from its more efficient Global-Aware Feature Decoupling. By directly analyzing the interplay between a sample's features and the entire set of global classification weights, our method is inherently more effective at identifying the nuanced discrepancies characteristic of near-OOD samples. This design provides a more robust and practical solution, striking a superior balance between high-precision detection and computational efficiency.

\begin{table*}[t]
\centering
\resizebox{.80\textwidth}{!}{
\begin{tabular}{lcccccccccc}
\toprule
\multirow{2}{*}{Methods}  & \multicolumn{2}{c}{ViT} & \multicolumn{2}{c}{Swin}  & \multicolumn{2}{c}{ResNet} & \multicolumn{2}{c}{DenseNet} \\
\cmidrule(lr){2-3}\cmidrule(lr){4-5}\cmidrule(lr){6-7}\cmidrule(lr){8-9}
& AUROC$\uparrow$ & FPR95$\downarrow$ & AUROC$\uparrow$ & FPR95$\downarrow$   & AUROC$\uparrow$ & FPR95$\downarrow$ & AUROC$\uparrow$ & FPR95$\downarrow$ \\
\midrule

MSP  & 79.34 & 66.29 & 77.81 & 67.14   & 73.92 & 70.73 & 77.77 & 67.64 \\
MaxLogit & 75.68 & 65.44 & 70.86 & 67.69 & 79.65 & 65.02 & 81.16 & 61.58 \\
ODIN & 61.46 & 94.34 & 55.22 & 91.02 & 79.28 & 60.35 & 78.91 & 60.65 \\
ReAct & 79.66 & 70.00 & 81.92 &	66.58  & 86.04 & 44.13 & 78.47 & 64.19 \\
Energy & 71.04 & 70.24 & 63.19 & 75.96  & 79.78 & 64.80 & 81.10 & 61.26 \\
ViM  & 86.72 & \textbf{49.38} & 83.97 & 63.74   & 83.97 & 65.88 & 81.34 & 72.67 \\
GEN & 85.36 & 59.53 & 83.73 & \textbf{55.28} & 79.94 & 66.97 & 82.40 & 63.01 \\
OptFS & 83.91 & 66.01 & 84.73 & 65.02 & 88.05 & 42.80 & \underline{88.01} & \underline{48.71} \\
CADRef  & \underline{86.91} & 60.07 & \underline{87.10} & 57.39  &  \underline{89.27} & \underline{40.69} & \textbf{88.64} & \textbf{45.24}\\
\midrule
\textbf{GAFD-CC(Ours)} & \textbf{87.05} & \underline{59.15} & \textbf{87.28} & \underline{56.08}  & \textbf{89.93} & \textbf{39.66} & 87.63 & \underline{48.71}\\
\bottomrule
\end{tabular}
}
\caption{Results of OOD detection on ImageNet-1K benchmark with various backbone models. This table presents the Average results for different OOD detection methods across various backbone architectures. The results demonstrate the adaptability and consistent performance of GAFD-CC (Ours) with diverse model backbones.}
\label{tab:more_models}
\end{table*}

\subsubsection{Cross-model robustness discussion}

GAFD-CC demonstrates remarkable and consistent performance across a diverse range of model architectures, including both Transformer-based (ViT-B/16 (\cite{dosovitskiy2021imageworth16x16words}), Swin-B (\cite{liu2021swintransformerhierarchicalvision}) and CNN-based (ResNet-50(\cite{he2015deepresiduallearningimage}), DenseNet-201 (\cite{huang2018denselyconnectedconvolutionalnetworks}) models, as detailed in Table~\ref{tab:more_models}. The pre-trained model weights are sourced from the PyTorch model zoo (\cite{NEURIPS2019_bdbca288}). Our method consistently achieves leading or highly competitive results in both AUROC and FPR95 metrics across these varied backbones, and notably secures the best overall performance.

For instance, on the ViT-B/16 architecture, GAFD-CC records the best AUROC of 87.05\%. On Swin-B, it secures the best AUROC (87.28\%) and the second best FPR95 (56.08\%). Among CNN-based models, GAFD-CC likewise leads on ResNet-50 with the best AUROC (89.93\%) and FPR95 (39.66\%), and on DenseNet-201 it achieves 87.63\% AUROC and 48.71\% FPR95, performing near-optimally.

This strong cross-model robustness can be attributed to GAFD-CC's fundamental design principles. Unlike methods that might rely on architecture-specific feature characteristics or complex, learned transformations (e.g., ViM's reliance on projecting features onto a specific ID subspace to extract residuals, or CADRef's focus on the single maximum-logit direction), GAFD-CC employs a global-aware feature decoupling mechanism. By considering the interplay between the entire feature vector and the complete set of classification weights, GAFD-CC extracts a more inherent and universal OOD signal that is less sensitive to the specific way different neural network architectures generate or represent features. Furthermore, its adaptive confidence calibration effectively integrates this decoupled feature information with robust logit-based scores, allowing for a flexible combination of signals that enhances generalizability. This inherent simplicity and generality in design enable GAFD-CC to generalize exceptionally well, validating its broad applicability and effectiveness irrespective of the underlying neural network architecture.

\begin{table}[t]
\centering
\resizebox{.65\columnwidth}{!}{
\begin{tabular}{lcc}
\toprule
Methods & AUROC$\uparrow$ & FPR95$\downarrow$ \\
\midrule
Base & 86.86 & 60.49 \\
Base + FD & 86.85 & 60.66 \\
Base + FD + CC & \textbf{87.05} & \textbf{59.15}\\
\bottomrule
\end{tabular}
}
\caption{Ablation studies on ImageNet-1K benchmark. All values are averaged across multiple OOD datasets.}
\label{tab:ablation}
\end{table}

\subsection{Ablation study} 
We conduct an ablation study to validate the individual and synergistic contributions of the key components in the GAFD-CC: Global-Aware Feature Decoupling (FD) and Confidence Calibration (CC). The results on the ImageNet-1K benchmark, utilizing a ViT-B/16 backbone, are presented in Table~\ref{tab:ablation}.

Our “Base” method for this study is an initial feature deviation score, representing a fundamental approach without the full GAFD-CC enhancements. As shown in Table~\ref{tab:ablation}, applying only the feature decoupling (Base + FD) to this base method results in a negligible or even slightly negative change in performance (AUROC from 86.86\% to 86.85\%, FPR95 from 60.49\% to 60.66\%), indicating that feature decoupling alone, without proper confidence calibration, is insufficient to enhance detection significantly. However, when we introduce the confidence calibration module alongside feature decoupling (Base + FD + CC), the performance improves substantially across both metrics. The AUROC increases to 87.05\% and FPR95 drops significantly to 59.15\%. This clearly demonstrates the powerful synergistic effect of our two proposed components: the Global-Aware Feature Decoupling provides more meaningful and discriminative feature components, and the Confidence Calibration effectively fuses these with multi-scale logit information to achieve superior and robust OOD detection performance.

\begin{figure}
\centering
\includegraphics[width=0.9\columnwidth]{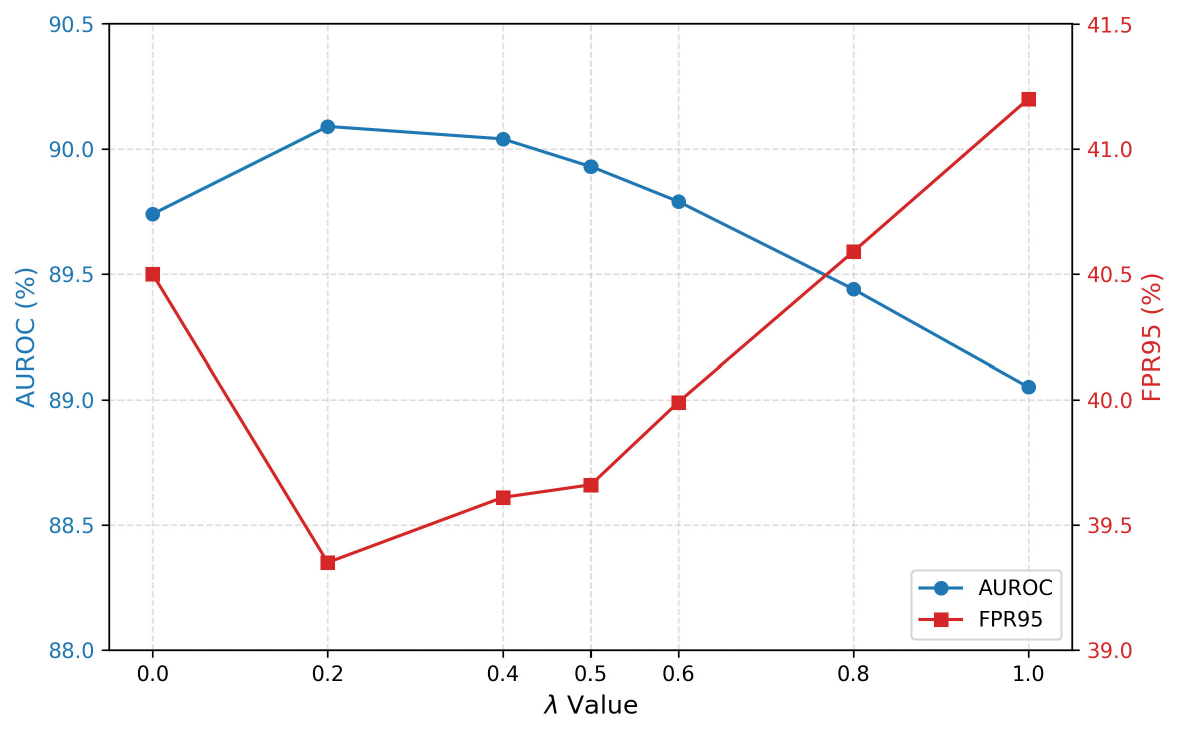}
\caption{ The impact of the hyperparameter $\lambda$ on OOD detection performance. This figure investigates how varying the hyperparameter $\lambda$, which governs the relative contribution of the two main components of the GAFD-CC score, affects the AUROC and FPR95 on test results.}
\label{fig:param}
\end{figure}

\subsection{Hyperparameter discussion} 

\subsubsection{Hyperparameter in feature decoupling} 

As indicated in the equation: 
\begin{equation}
\begin{aligned}
Score(x) = \lambda\frac{\xi^{+}(x)}{s_{sample} + s_{class}} + (1 -\lambda)
\frac{\xi^{-}(x)}{s_{global}+s_{class}}.
\end{aligned}
\end{equation}
We introduce the positive-feature coefficient $\lambda$ to investigate how the relative weighting of the two main components affects detection performance. As shown in Figure~\ref{fig:param}, the FPR95 curve first decreases and then increases as $\lambda$ grows, reaching its best value around $\lambda = 0.2$. This trend confirms the key role of the positive-feature term: it refines the ID/OOD decision boundary by improving the separability of hard samples near that boundary, yielding a more robust threshold. Nevertheless, when the positive-feature term becomes too dominant, the AUROC curve also rises first and then falls, again peaking near $\lambda = 0.2$. Excessive weight on positive features injects noise that enlarges intra-class variance and ultimately degrades ranking quality. Conversely, the negative-feature term characterizes atypical patterns for ID data and provides counter-evidence that pulls the decision boundary inward; this suppresses false positives and tightens the ID distribution. Moreover, it acts as an important regularizer when the positive term dominates, restraining intra-class variance and helping to stabilize AUROC. 

Although Figure~\ref{fig:param} indicates that when the negative-feature-related term is dominant ($\lambda=0.2$), OOD detection performance reaches its optimum, the performance curve shows little overall variation around the 1:1 ratio ($\lambda = 0.5$). Therefore, for the sake of simplicity and generality, choosing the 1:1 ratio ($\lambda = 0.5$) is a reasonable compromise, providing near-optimal robust performance without requiring additional hyperparameter tuning.

\begin{figure}
\centering
\includegraphics[width=0.9\columnwidth]{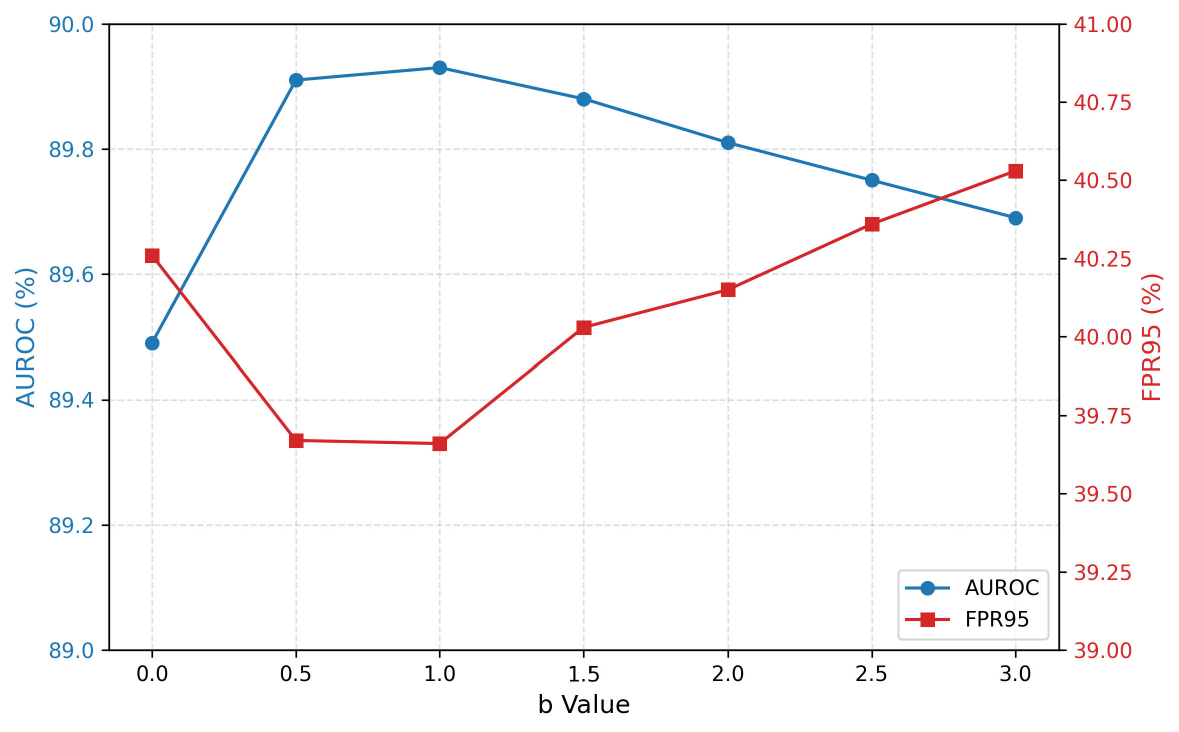}
\caption{ The impact of the hyperparameter $b$ on OOD detection performance. This figure investigates how varying the hyperparameter $b$, which governs the influence of class-related confidence on detection performance.}
\label{fig:paramb}
\end{figure}

\subsubsection{Hyperparameter in confidence calibration} 

As indicated in the equation: 
\begin{equation}
\begin{aligned}
Score(x) = \frac{\xi^{+}(x)}{s_{sample} + b \cdot s_{class}} + \frac{\xi^{-}(x)}{s_{global}+ b \cdot s_{class}}.
\end{aligned}
\end{equation}
We introduce $b$ to investigate the impact of class-mean confidence on detection performance. As shown in Table~\ref{tab:ood_detection_resnet}, GAFD-CC($b=1$) exhibits superior performance across multiple datasets and achieves the highest average score, demonstrating greater robustness. Conversely, GAFD-C($b=0$) performs better on OOD data exhibiting specific distributional shifts relative to the In-Distribution (ID) dataset, such as iNaturalist.

We further analyze how varying $b$ affects detection performance, which modulates the influence of class-mean confidence in score refinement. As shown in Figure~\ref{fig:paramb}, When $b = 0$, only sample and global confidence are retained; discarding the class prior results in insufficient calibration at the distribution tails and a consequently higher FPR95. Conversely, when$ b > 1$, class-mean confidence becomes excessively dominant, which amplifies the inherent bias of the class prior and introduces noise, particularly degrading AUROC on class-imbalanced datasets like iNaturalist and SUN. Optimally, when $b = 1$, the three confidence sources are fused with equal weights. This approach successfully supplements class information while mitigating overfitting to any single metric, thus achieving an optimal balance without requiring additional hyperparameter search.

\section{Conclusion}
In this paper, we introduced a novel method for OOD detection, GAFD-CC. It enhances the detection performance for out-of-distribution samples through global class-aware feature decoupling and logit-based score confidence fusion. Extensive experiments on various benchmarks demonstrate GAFD-CC's effectiveness and robustness. Furthermore, comprehensive evaluations across multiple CNN- and Transformer-based architectures further validate its efficacy and cross-model robustness. Ablation studies further confirm the effectiveness of each component within the GAFD-CC scoring framework. In future work, we will explore advanced hybrid strategies to improve OOD detection performance and reliability across diverse datasets and model architectures.

\printcredits

\section*{Declaration of competing interest}

The authors declare that they have no known competing financial interests or personal relationships that could have appeared to influence the work reported in this paper.

\section*{Data availability}
All data generated or analyzed during this study are included in this 
published article.

\bibliographystyle{cas-model2-names}

\bibliography{cas-refs}


\end{document}